\documentclass{article}

\usepackage{arxiv}
\usepackage{amsmath,amsfonts}
\usepackage{algorithmic}
\usepackage{array}
\usepackage[caption=false,font=normalsize,labelfont=sf,textfont=sf]{subfig}
\usepackage{textcomp}
\usepackage{stfloats}
\usepackage{url}
\usepackage{verbatim}
\usepackage{multirow}

\pdfoutput=1

\usepackage{times}
\usepackage{latexsym}

\usepackage[T1]{fontenc}

\usepackage[utf8]{inputenc}

\usepackage{microtype}

\usepackage{inconsolata}
\usepackage{multirow}
\usepackage{graphicx}

\title{ASR-EC Benchmark: Evaluating Large Language Models on Chinese ASR Error Correction}

\author{Victor Junqiu Wei, Weicheng Wang \\
Department of Computer Science and Engineering\\
The Hong Kong University of Science and Technology\\
\texttt{wjqjsnj@gmail.com, wwangby@connect.ust.hk} \\
\And
Di Jiang, Yuanfeng Song\\
AI Group, WeBank Co., Ltd, Shenzhen, China\\
\texttt{\{dijiang,yfsong\}@webank.com}
\And
Lu Wang\\
Shenzhen University, China\\
\texttt{wanglu@szu.edu.cn}
}

\begin{document}
\maketitle
\begin{abstract}
Automatic speech Recognition (ASR) is a fundamental and important task in the field of speech and natural language processing. It is an inherent building block in many applications such as voice assistant, speech translation, etc. Despite the advancement of ASR technologies in recent years, it is still inevitable for modern ASR systems to have a substantial number of erroneous recognition due to environmental noise, ambiguity, etc. Therefore, the error correction in ASR is crucial. 

Motivated by this, this paper studies ASR error correction in the Chinese language, which is one of the most popular languages and enjoys a large number of users in the world. We first create a benchmark dataset named \emph{ASR-EC} that contains a wide spectrum of ASR errors generated by industry-grade ASR systems. To the best of our knowledge, it is the first Chinese ASR error correction benchmark. Then, inspired by the recent advances in \emph{large language models (LLMs)}, we investigate how to harness the power of LLMs to correct ASR errors. We apply LLMs to ASR error correction in three paradigms. The first paradigm is prompting, which is further categorized as zero-shot, few-shot, and multi-step. The second paradigm is finetuning, which finetunes LLMs with ASR error correction data. The third paradigm is multi-modal augmentation, which collectively utilizes the audio and ASR transcripts for error correction. Extensive experiments reveal that prompting is not effective for ASR error correction. Finetuning is effective only for a portion of LLMs. Multi-modal augmentation is the most effective method for error correction and achieves state-of-the-art performance.
\end{abstract}

\section{Introduction}

\emph{Automatic Speech Recognition (ASR)} refers to the technology that enables computers to recognize and interpret human speech, converting it into text. 
It finds wide applications in voice assistants, speech dialogue systems, speech translations, etc. 
Although this technology has advanced significantly in recent years, it is still inevitable for modern ASR systems to have erroneous recognition due to environmental noise, ambiguity, etc. 
Thus, the ASR error correction is an important problem for the speech and language processing. 

In this paper, we study the ASR error correction for the Chinese language. 
Chinese is one of the most popular languages in the world and enjoys a large number of users. 
Despite the popularity of Chinese, we observe that there are no existing ASR error correction datasets for the Chinese language. 

Motivated by this, we establish a benchmark dataset for Chinese ASR error correction \cite{OurData2024}. Based upon the open-source ASR toolkit Kaldi-K1\cite{Povey_ASRU2011} and Kaldi-K2\footnote{https://kaldi-asr.org/}\footnote{https://github.com/k2-fsa/k2}, we construct the ASR-EC benchmark by processing audio clips from THCHS-30, AISHELL-1, AISHELL-2, and WeNetSpeech. 
This dataset encapsulates a broad range of decoding errors and is designed to assess LLMs' capability to correct ASR mistakes across varied utterance lengths.

We further investigate how to utilize LLMs~\cite{devlin-etal-2019-bert,brown2020language, zhao2023survey} for ASR error correction. 
There are three paradigms. 
We first prompt these models to act as an error correction module for existing Chinese ASR systems. 
Then, with parameter-efficient fine-tuning \cite{hu2021lora}, we customize the models to the context of the Chinese language and the ASR task. 
Finally, through multimodal augmentation, we leverage both the audio and text modalities to enhance the LLMs' understanding of the content, providing a comprehensive basis for the LLMs to detect and correct errors. 

Our experiments demonstrate that different strategies for applying LLMs to ASR error correction yield various degrees of effectiveness. 
Prompting is to correct errors by simply querying foundation models with the erroneous text. 
This method has proven to be ineffective and can even introduce new errors to previously correct content. 
This implies that, without annotated ASR error correction datasets, 
LLMs cannot achieve satisfactory performance even if the advanced prompting method is applied. 
In comparison, finetuning enables models to leverage their contextual understanding and language mastery to meaningfully refine the ASR output, correcting various decoding mistakes. 
Moreover, multimodal augmentation stands out as the most effective approach, significantly enhancing error correction by jointly analyzing the audio and its corresponding transcript, thereby achieving state-of-the-art performance in correcting ASR errors.

The contributions of this paper are threefold:
\begin{itemize}
    \item We build and release a public dataset named ASR-EC for LLM-based ASR error correction. To the best of our knowledge, this is the first dataset in the Chinese ASR error correction. 
    This benchmark will pave the way for future studies on the Chinese ASR error correction. 
    \item We undertake a comprehensive investigation of three paradigms for adapting LLMs to ASR error correction, namely \emph{prompting}, \emph{funetuning}, and \emph{multi-modal}. 
    \item We conducted an empirical study on these LLM-based paradigms for ASR error correction on our constructed benchmark. We found that multi-modal augmentation stands out as the best approach. 
\end{itemize}

The discovery in this paper represents a promising direction to inject powerful LLMs into conventional ASR pipelines and significantly improve their performance. 
We will release our datasets and source code upon the publication of this paper. 

The remainder of this paper is organized as follows. 
Section~\ref{sec:related} reviews the related work of error correction. 
Section~\ref{sec:benchmark} demonstrates the construction of our proposed ASR-EC benchmark for Chinese ASR error correction. 
Sections~\ref{sec:prompt}, \ref{sec:finetune}, and \ref{sec:mulitmodal} present our investigated three approaches for LLM-based Chinese ASR error correction. 
Section~\ref{sec:exp} presents our empirical study, and finally, Section~\ref{sec:con} concludes this paper. 

\begin{table*}[!h]
\centering

\begin{tabular}{c|c|c|c|c|c}
\hline 
Corpus & Source & Transcription & \multicolumn{1}{l|}{\# Hours} & \multicolumn{1}{l|}{\# Utterances} & Avg Characters \\ \hline \hline
THCHS-30 & Recorded speech & Manually labeled & 30 & 13,388 & 33   \\ \hline
AISHELL-1 & Recorded speech & Manually labeled & 200 & 141,597 & 14  \\ \hline
AISHELL-2 & Recorded speech & Manually labeled & 1,000 & 511,123 & 13   \\ \hline
WeNetSpeech & YouTube, Podcast & OCR, ASR & 1,100 & 434,781 & 38  \\ \hline
\end{tabular}%
\vspace{0.4cm}
\resizebox{\columnwidth}{!}{%

\begin{tabular}{c|c}
\hline 
Corpus  & Domains  \\ \hline \hline
THCHS-30 &  News \\ \hline
AISHELL-1 & Smart home, autonomous driving, and  industrial production, etc. \\ \hline
AISHELL-2 & Keywords, voice command, smart home,  autonomous driving, industrial production, etc. \\ \hline
WeNetSpeech &  Audiobook, commentary, documentary,  drama, interview, etc. \\ \hline
\end{tabular}%
}
\caption{Speech Corpora Statistics}
\label{tab:profile}
\end{table*}

\begin{figure*}
    \centering
    \includegraphics[width=\textwidth]{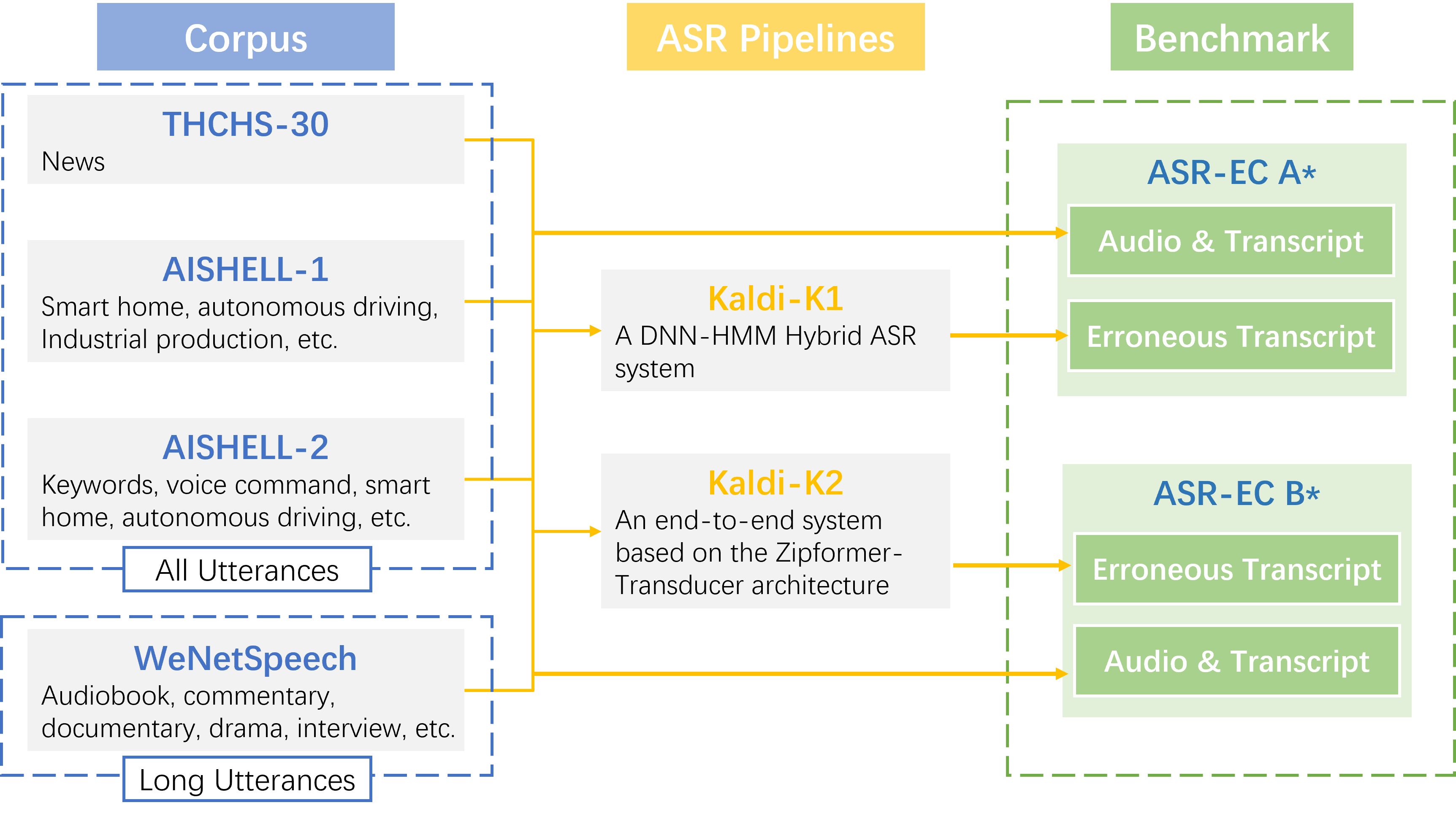}
    \caption{Pipelines for Erroneous ASR Transcripts Construction}
    \label{fig:enter-label}
\end{figure*}

\section{Related Work}
\label{sec:related}
Error correction models, proposing to identify and correct inaccuracies in the text and audio, play a crucial role in Automatic Speech Recognition (ASR). 
Their development mirrored the advancements in Natural Language Processing (NLP). 

\smallskip\noindent\textbf{Text Error Correction.} 
Initially, \emph{rule-based} models were predominant in error correction. 
These models relied on predefined rules and heuristics to correct text errors, which often limited their adaptability.
The advent of \emph{end-to-end} models marks a large leap forward~\cite{hrinchuk2020correction, zhao2021bart, jiang2023probabilistic}. 
Instead of requiring manually defined rules, they can learn directly from data. 
This adaptability leads to higher accuracy and more effective error correction. 

\smallskip\noindent\textbf{Text and Audio Error Correction.} LLMs have shown considerable potential in error correction. 

Firstly, for the text,~\cite{ma2023n, Yang_2023} prompted and fine-tuned LLMs with ASR error correction data, transferring the knowledge learned by the large-scale pre-trained language model from vast textual data to error correction tasks. 
\cite{ma2023can} studied the error correction performance of the most advanced Large Language Model (LLM) at present, ChatGPT. 
\cite{hu2024large} extended generative error correction benchmarks to noisy conditions,
showcasing LLMs' dual capability in denoising and error correction. 

Secondly, for the multimodal contexts, including audio and text, the Qwen-Audio model~\cite{chu2023qwenaudio} advances audio-language understanding by pre-training across various tasks and audio types, leading to a versatile model that enhances multi-turn dialogues and audio-centric interactions without needing fine-tuning. 
\cite{chen2024its} integrates acoustic data into the generative error correction process, enhancing the model's ability to map from N-best ASR hypotheses to accurate transcriptions.

\textbf{Language Model and Large Language Model (LLM). }The language model (LM) is a crucial component of ASR systems and it helps transform the output of the acoustic model (AM) into coherent and natural sentences. With advancements in natural language processing, language models have seen significant performance improvements, resulting in enhanced ASR performance.

The primary language model is the bag-of-words model (BoW), which represents text in a one-hot format. This method is suitable for processing discrete data and extending features, but it does not take into account the order of words \cite{baeza2015predicting}. While this approach is straightforward, its semantic representation lacks precision. In ASR, the most commonly used language model is the N-gram model, which estimates the probability of the next word by counting co-occurrences. Despite its simplicity, the robustness and interpretability of the N-gram model make it a valuable choice for language modeling in ASR systems.

In contrast to statistical language models, Bengio et al. introduced the concept of neural network language models in 2003 \cite{bengio2003neural}. The performance of language models has significantly improved with the advent of word vector models. The most notable examples are Word2Vec \cite{mikolov2013efficient} and GloVe \cite{pennington2014glove}, which convert each word into a vector representation that captures richer semantic information. Since contextual words provide valuable insights, researchers \cite{wang2018densely} have proposed using convolutional neural networks (CNNs) instead of N-grams to capture context from a larger receptive field. Additionally, the success of recurrent neural network language models (RNNLMs) underscores the importance of modeling long-range sequential context.

On the other hand, pre-training models have become the dominant approach in language modeling \cite{jiang2023probabilistic, lu2024pretraining,wu2022phonetic,zhou2018multi,devlin2018bert,radford2019language,lan2019albert,liu2019roberta,wei2021training,wei2019nezha,li2021heterogeneous,jiang2021industrial,zhou2021memetic,jiang2019federated, song2020topicocean,hong2024expanding,jiang2016latent}. ELMo \cite{peters2018deep} utilizes an LSTM architecture with bidirectional settings to fully capture contextual information, reflecting the characteristics of different dimensions. Pre-trained models can handle various downstream tasks and generally exhibit superior performance. The Transformer architecture \cite{vaswani2017attention} (and its variants \cite{zhang2021continuous,zhang2020tensorcoder,wang2022clusterformer,li2022hypoformer}), known for its powerful attention mechanism, has demonstrated exceptional capability in modeling deep structural information from data \cite{radford2018improving,zhou2018multi}.

Transformers have quickly been adopted in large-scale pre-trained language models, including GPT \cite{radford2018improving}, BERT \cite{devlin2018bert}, GPT-2 \cite{radford2019language}, and subsequent models like ALBERT \cite{lan2019albert} and RoBERTa \cite{liu2019roberta}. However, it remains an open question how to effectively utilize and integrate language models in ASR correction.

To the best of our knowledge, there are no existing benchmarks for Chinese ASR error correction, even though Chinese is one of the most popular languages and has a large number of users in the world. 
Motivated by this, this paper constructs the first Chinese ASR error correction benchmark and paves the way for future research on the ASR error correction. 
In addition, we systematically investigate the paradigms for the LLM-based ASR error correction, i.e., prompting-based, finetuning-based, and multi-modal paradigms. 
Our results demonstrate that different paradigms yield various degrees of effectiveness and the multi-modal approach stands out as the best one. 
As far as we are concerned, our research work is the first one to study LLM-based ASR error correction and systematically study the paradigms of ASR error correction with LLMs.

\section{ASR-EC Benchmark}
\label{sec:benchmark}

\subsection{Speech Corpora Collection}

To build the ASR-EC benchmark, we curate a collection from four open-source Chinese speech corpora: THCHS-30~\cite{wang2015thchs}, AISHELL-1~\cite{bu2017aishell}, AISHELL-2~\cite{du2018aishell}, and WeNetSpeech~\cite{zhang2022wenetspeech}. The characteristics of these corpora are detailed in Table~\ref{tab:profile}. 

Our study leverages the full datasets of AISHELL-1, AISHELL-2, and THCHS-30. Notably, these three datasets contain a relatively low percentage of long utterances, and they are all one-sentence recognition tasks. 
In order to evaluate the error correction capabilities of LLMs on long utterances, we also select 434,781 audio clips whose transcripts contain more than 30 Chinese characters. 
Instead, WeNetSpeech contains both one-sentence recognition tasks and multi-sentence recognition tasks. 
Thus, WeNetSpeech contains a rather higher percentage of long utterances and is supposed to be harder than the other three datasets.

\subsection{ASR-EC Benchmark}

We adopt two well-recognized ASR pipelines, namely \emph{Kaldi-K1} and \emph{Kaldi-K2}, to establish the erroneous transcripts. 
To the best of our knowledge, they are the only two pipelines in the history of ASR systems. 
These two pipelines have contrasting architectures and different ASR performance. 
In this paper, we adopt both pipelines to create a large variety of different types of errors and different levels of difficulties in our benchmark. 
Kaldi-K1~\cite{Povey_ASRU2011}, a DNN-HMM Hybrid ASR system, employs a multistream CNN for acoustic modeling and an n-gram language model formatted by FST structure. 
Kaldi-K2, an end-to-end system, is based on the Zipformer-Transducer architecture. Both of the two ASR pipelines are pre-trained on a 10000-hour Chinese speech corpus. 

Based on the decoding results of Kaldi-K1 and Kaldi-K2, we establish two datasets in the ASR-EC. In order to reflect LLMs' performance on utterances of different lengths, we divided each dataset into two subsets: short utterances and long utterances. The average utterance length in the short utterance subset is about 13 characters, while in the long utterance subset, it is about 38 characters. Table \ref{tab:statistics} reports the statistics of these datasets and subsets. To control the number of those with a CER of 0, only 10\% of them are kept. 
Besides, in this table, we also consider three types of ASR errors, namely \emph{substitution}, \emph{deletion} and \emph{insertion}. 
The percentages of the three types of error in the short, long and mixed utterances are also summarized in the table. 

\begin{table*}[htp]
\centering
\resizebox{\columnwidth}{!}{%
\begin{tabular}{lll|r|r|r}
\hline
 &  &  & \multicolumn{1}{l|}{\begin{tabular}[c]{@{}l@{}}Short Utter Subset\end{tabular}} & \multicolumn{1}{l|}{\begin{tabular}[c]{@{}l@{}}Long Utter Subset\end{tabular}} & \multicolumn{1}{l}{\begin{tabular}[c]{@{}l@{}}Whole Dataset\end{tabular}} \\ \hline
\multicolumn{1}{c|}{\multirow{ 4}{*}{ASR-EC A*}} & \multicolumn{1}{l|}{\multirow{2}{*}{Train Set}} & CER(\%) & 13.49 & 12.24 & 12.45 \\
\multicolumn{1}{c|}{} & \multicolumn{1}{l|}{} & \#Utter & 186,150 & 358,400 & 544,551 \\ \cline{2-6} 
\multicolumn{1}{c|}{} & \multicolumn{1}{l|}{\multirow{2}{*}{Test Set}} & CER(\%) & 12.83 & 11.73 & 12.42 \\
\multicolumn{1}{c|}{} & \multicolumn{1}{l|}{} & \#Utter & 1,024 & 1,024 & 1,024 \\ \hline
\multicolumn{1}{l|}{\multirow{ 4}{*}{ASR-EC B*}} & \multicolumn{1}{l|}{\multirow{2}{*}{Train Set}} & CER(\%) & 12.99 & 7.03 & 8.15 \\
\multicolumn{1}{l|}{} & \multicolumn{1}{l|}{} & \#Utter & 181,488 & 280,520 & 462,009 \\ \cline{2-6} 
\multicolumn{1}{l|}{} & \multicolumn{1}{l|}{\multirow{2}{*}{Test Set}} & CER(\%) & 13.44 & 6.92 & 8.11 \\
\multicolumn{1}{l|}{} & \multicolumn{1}{l|}{} & \#Utter & 1,024 & 1,024 & 1,024 \\ \hline
\end{tabular}%
}
\vspace{0.5cm}
\resizebox{\columnwidth}{!}{%
\begin{tabular}{l|c|r|r|r}
\hline
 &   & \multicolumn{1}{l|}{\begin{tabular}[c]{@{}l@{}}Short Utter Subset\end{tabular}} & \multicolumn{1}{l|}{\begin{tabular}[c]{@{}l@{}}Long Utter Subset\end{tabular}} & \multicolumn{1}{l}{\begin{tabular}[c]{@{}l@{}}Whole Dataset\end{tabular}} \\ \hline
{\multirow{ 4}{*}{ASR-EC A*}} & Substitution CER(\%)  & 10.54 & 6.13 & 8.21 \\\cline{2-5} 
 & Deletion CER(\%) & 2.20 & 3.04 & 2.81 \\ \cline{2-5} 
 & Insertion CER(\%) & 1.34 & 2.74 & 2.30 \\ \cline{2-5} 
  & Overall CER(\%) & 14.09 & 11.91 & 13.32 \\ \hline 
{\multirow{ 4}{*}{ASR-EC B*}} & Substitution CER(\%) & 9,59 & 2.60 & 5.94 \\\cline{2-5} 
& Deletion CER(\%) & 4.17 & 1.01 & 1.95 \\ \cline{2-5} 
 & Insertion CER(\%) & 1.57 & 3.46 & 2.50 \\ \cline{2-5} 
 & Overall CER(\%) & 15.33 & 7.08 & 10.39 \\ \hline
\end{tabular}%
}

\caption{Benchmark CER Results (ASR-EC A*, and B* are respectively generated by Kaldi-K1, a DNN-HMM-Hybrid ASR system,  and Kaldi-K2, an end-to-end system which is based on the Zipformer-Transducer architecture)}
\label{tab:statistics}
\end{table*}

\section{Prompting-based Correction}
\label{sec:prompt}
Prompting is an emerging technique for fine-tuning large language models (LLMs) in a parameter-efficient way. The prompts provide soft cues to steer the model behavior towards desired tasks, without modifying the actual parameters. We introduce two prompting methods, namely direct prompting and multi-step prompting, for conducting ASR correction using LLMs. Direct prompting encompasses both zero-shot and few-shot error correction techniques.

\textbf{Zero-shot prompting.} In zero-shot prompting, a prompt is presented to the model without accompanying explicit examples. The model is then anticipated to generate a response utilizing its pre-existing knowledge base. This method is suitable for quick and simple tasks.

\textbf{Few-shot prompting.} In few-shot prompting, we provide the model with three input-output pairs as examples, enabling it to learn from this isolated input. 
In each input-output pair, the input is the raw ASR translated text, which may or may not contain errors, and the output is the correct transcripts without any mistakes. 
This technique is beneficial for ensuring that the model's output follows a specific and consistent format.

\textbf{Multi-step prompting.} Given that LLMs lack context or prior knowledge about the errors in the outputs of ASR systems, correcting these errors using only direct prompting can be challenging. 
Multi-step prompting is a technique that breaks down a complex problem into smaller, manageable steps. 
LLMs process each of these steps sequentially to achieve the final objective. 
In particular, we employ a two-step prompting strategy, where the first step detects if the ASR output contains errors or not. 
After that, the second step will correct the errors if there are any errors detected in the first step.
Otherwise, the second step simply outputs the ASR transcript.

\section{Finetuning-based Correction}
\label{sec:finetune}

By prompting LLMs to perform ASR error correction tasks, LLMs lack a deep understanding of the task. 
They are unaware of the possible error types in the hypothesis texts generated by ASR and the corresponding ways of correction. 
Besides, LLMs may struggle with precisely following instructions, leading to challenges in accurately executing ASR error correction tasks.  

Fine-tuning LLMs is beneficial for making them more adaptable to downstream tasks and ensuring their outputs align with expectations. 
To enable LLMs to understand the task pattern and expected output of ASR error correction, we fine-tune selected LLMs in this paper. 
Note that since full fine-tuning is time-consuming and costly, we opt for a parameter-efficient fine-tuning method. Specifically, we use the LoRA fine-tuning method, which is a breakthrough and efficient fine-tuning technique. LoRA freezes the pre-trained model parameters and injects trainable rank decomposition matrices into each layer of the Transformer architecture. This significantly cuts down the number of trainable parameters.

\section{Correction based on Multimodal Augmentation}
\label{sec:mulitmodal}
\begin{figure}[ht]
  \centering
  \includegraphics[width=0.6\linewidth]{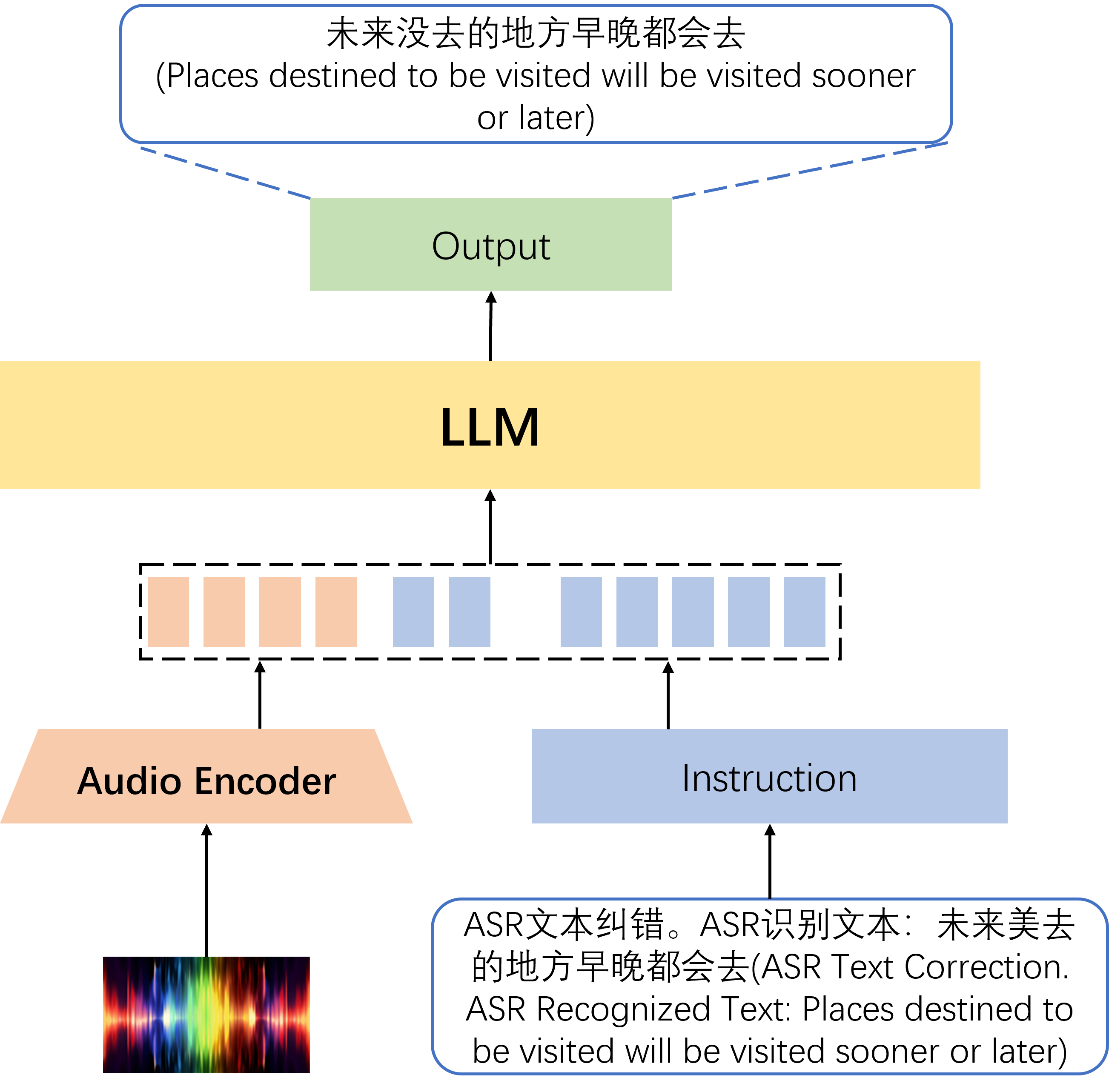}
  \caption{ASR error correction with Multimodal model.}
  \label{fig:model_output1}
\end{figure}

The incorporation of multimodal augmentation into ASR error correction leverages the synergy between audio and text modalities, offering a nuanced approach to identifying and correcting speech recognition errors. 

Multimodal augmentation employs a dynamic fusion process that integrates the complementary strengths of audio signals and textual data. This fusion enables a comprehensive understanding of content, allowing for the detection and correction of errors that may not be apparent when analyzing either modality in isolation. Once errors are identified, multimodal augmentation applies contextually informed corrections, ensuring that the final text accurately reflects the intended speech content. 
Figure~\ref{fig:model_output1} demonstrates the architecture of our multi-modal augmentation approach. 
In the input side of the LLM, we concatenate the encoded raw audio input and the instruction which includes the error correction prompt (e.g., ``ASR Text Correction, ASR Recognized Text: '') and the erroneous ASR text. The output text of LLM is the corrected text. 
The multimodal LLM is trained with our proposed dataset in an end-to-end fashion. 
This method proves especially effective in scenarios where traditional single-modality error correction techniques may fall short.

\section{Experiment}
\label{sec:exp}
\subsection{LLM Testbeds}
In order to demonstrate the effectiveness of the benchmark dataset ASR-EC, we evaluate three open-source large language models that support the Chinese language: ChatGLM3~\cite{du2022glm}, Qwen~\cite{bai2023qwen}, and Baichuan2~\cite{yang2023baichuan}. 
The information of these LLMs are shown in Table~\ref{tab:models}. 

Qwen-Audio\cite{chu2023qwenaudio} (Qwen Large Audio Language Model) is the multimodal extension of the expansive Qwen model series, also known as Tongyi Qianwen, introduced by Alibaba Cloud. Qwen-Audio is designed to process a wide range of audio inputs, including human speech, natural sounds, music, and songs, along with textual inputs. It then generates textual outputs based on the given inputs.

\begin{table}[htp]
    \centering
    \begin{tabular}{c|c|c}\hline
        Model & No. of Parameters & Modality \\ \hline \hline
        ChatGLM3 & 6B & Text \\ \hline
        Qwen & 7B & Text \\ \hline
        Baichuan2 & 7B & Text \\ \hline
        Qwen-Audio & 7B & Text \& Audio \\ \hline
    \end{tabular}
    \caption{Information of LLMs}
    \label{tab:models}
\end{table}

During model inference, we set the generation configuration of Qwen~\cite{bai2023qwen} and Baichuan2~\cite{yang2023baichuan} models to their default value. When using ChatGLM3 model for inference, it was observed that the default parameters would cause the model to generate repetitive text. In order to reduce the repetition of the model output, we adjusted the repetition penalty parameter to 1.05 (the default value is 1), and the rest of the parameters remain the default~\cite{du2022glm}. When conducting LoRA fine-tuning, we referenced the fine-tuning hyperparameters from both the official repositories of these models and some third-party repositories~\cite{db-gpt-hub}\cite{llama-factory}. 

In this experiment, we compare our approaches with ASR baselines. For the dataset ASR-EC $A^*$, we compare ours with Kaldi-K1 which is the ASR model for generating the erroneous transcript in ASR-EC $A^*$. 
For the dataset ASR-EC $B^*$, we compare ours with Kaldi-K2 which is the ASR model for generating the erroneous transcript in ASR-EC $B^*$. 
We also compare with the state-of-the-art Chinese text error correction model called \emph{ReLM}~\cite{liu2024chinese}. 
In particular, for each erroneous transcript in ASR-EC $A^*$ and ASR-EC $B^*$, we adopt \emph{ReLM} to correct the errors and report the CER achieved.

\subsection{Prompting and Finetuning}

\begin{table*}[htp]
\centering
\begin{tabular}{cl|rrr|rrr}
\hline
 & \multicolumn{1}{c|}{} & \multicolumn{3}{c|}{CER of Test Set of ASR-EC A*} & \multicolumn{3}{c}{CER of Test Set of ASR-EC B*} \\ \cline{3-8} 
 & \multicolumn{1}{c|}{} & \multicolumn{1}{c|}{\begin{tabular}[c]{@{}c@{}}Short \\ Utterances\end{tabular}} & \multicolumn{1}{c|}{\begin{tabular}[c]{@{}c@{}}Long \\ Utterances\end{tabular}} & \multicolumn{1}{c|}{\begin{tabular}[c]{@{}c@{}}Mixed \\  Utterances\end{tabular}} & \multicolumn{1}{c|}{\begin{tabular}[c]{@{}c@{}}Short \\ Utterances\end{tabular}} & \multicolumn{1}{c|}{\begin{tabular}[c]{@{}c@{}}Long \\ Utterances\end{tabular}} & \multicolumn{1}{c}{\begin{tabular}[c]{@{}c@{}}Mixed\\  Utterance\end{tabular}} \\ \hline
\multicolumn{2}{c|}{ASR Baselines} & \multicolumn{1}{r|}{\textit{\textbf{12.83}}} & \multicolumn{1}{r|}{\textit{\textbf{11.73}}} & \textit{\textbf{12.42}} & \multicolumn{1}{r|}{\textit{\textbf{13.44}}} & \multicolumn{1}{r|}{\textit{\textbf{6.92}}} & \textit{\textbf{8.11}} \\ \hline
\multicolumn{1}{c|}{\multirow{ 3}{*}{Zero-Shot}} & Baichuan2 & \multicolumn{1}{r|}{27.47} & \multicolumn{1}{r|}{34.13} & {33.05} & \multicolumn{1}{r|}{26.81} & \multicolumn{1}{r|}{31.23} & {30.41} \\ \cline{2-8} 
\multicolumn{1}{c|}{} & ChatGLM3 & \multicolumn{1}{r|}{\textbf{21.92}} & \multicolumn{1}{r|}{\textbf{24.56}} & \textbf{24.13} & \multicolumn{1}{r|}{\textbf{20.99}} & \multicolumn{1}{r|}{\textbf{20.66}} & \textbf{20.73} \\ \cline{2-8} 
\multicolumn{1}{c|}{} & Qwen & \multicolumn{1}{r|}{27.21} & \multicolumn{1}{r|}{25.58} & {25.84} & \multicolumn{1}{r|}{26.04} & \multicolumn{1}{r|}{21.30} & {22.15} \\ \hline
\multicolumn{1}{c|}{\multirow{ 3}{*}{Three-Shot}} & Baichuan2 & \multicolumn{1}{r|}{22.54} & \multicolumn{1}{r|}{26.33} & {25.71} & \multicolumn{1}{r|}{22.04} & \multicolumn{1}{r|}{22.31} & {22.26} \\ \cline{2-8} 
\multicolumn{1}{c|}{} & ChatGLM3 & \multicolumn{1}{r|}{\textbf{19.67}} & \multicolumn{1}{r|}{\textbf{23.64}} & \textbf{22.97} & \multicolumn{1}{r|}{\textbf{18.87}} & \multicolumn{1}{r|}{\textbf{19.69}} & \textbf{19.53} \\ \cline{2-8} 
\multicolumn{1}{c|}{} & Qwen & \multicolumn{1}{r|}{21.36} & \multicolumn{1}{r|}{24.30} & {23.74} & \multicolumn{1}{r|}{20.44} & \multicolumn{1}{r|}{20.59} & {20.56} \\ \hline
\multicolumn{1}{c|}{\multirow{ 3}{*}{Multi-step}} & Baichuan2 & \multicolumn{1}{r|}{\textbf{22.23}} & \multicolumn{1}{r|}{25.45} & {24.92} & \multicolumn{1}{r|}{\textbf{21.57}} & \multicolumn{1}{r|}{21.44} & {21.46} \\ \cline{2-8} 
\multicolumn{1}{c|}{} & ChatGLM3 & \multicolumn{1}{r|}{23.12} & \multicolumn{1}{r|}{\textbf{23.02}} & \textbf{23.03} & \multicolumn{1}{r|}{22.27} & \multicolumn{1}{r|}{\textbf{18.65}} & \textbf{19.32} \\ \cline{2-8} 
\multicolumn{1}{c|}{} & Qwen & \multicolumn{1}{r|}{26.73} & \multicolumn{1}{r|}{27.26} & {27.12} & \multicolumn{1}{r|}{25.88} & \multicolumn{1}{r|}{22.68} & {23.27} \\ \hline
\multicolumn{1}{c|}{\multirow{3}{*}{\parbox{1.6cm}{LoRA Finetuning}}} & Baichuan2 & \multicolumn{1}{r|}{\textbf{10.99}} & \multicolumn{1}{r|}{\textbf{11.71}} & \textbf{12.36} & \multicolumn{1}{r|}{\textbf{11.21}} & \multicolumn{1}{r|}{\textbf{6.97}} & \textbf{7.88} \\ \cline{2-8} 
\multicolumn{1}{c|}{} & ChatGLM3 & \multicolumn{1}{r|}{13.10} & \multicolumn{1}{r|}{12.84} & 13.46 & \multicolumn{1}{r|}{12.96} & \multicolumn{1}{r|}{7.43} & 8.57 \\ \cline{2-8} 
\multicolumn{1}{c|}{} & Qwen & \multicolumn{1}{r|}{13.45} & \multicolumn{1}{r|}{14.16} & 14.59 & \multicolumn{1}{r|}{12.58} & \multicolumn{1}{r|}{7.08} & 8.48 \\ \hline
\multicolumn{1}{c|}{\multirow{3}{*}{\parbox{1.6cm}{Multimodal}}} &
Baichuan2 & \multicolumn{1}{r|}{\textit{\textbf{6.81}}} & \multicolumn{1}{r|}{{{6.99}}} & \textit{\textbf{5.96}} & \multicolumn{1}{r|}{\textit{\textbf{5.75}}} & \multicolumn{1}{r|}{\textit{\textbf{4.29}}} & \textit{\textbf{5.12}} \\
\cline{2-8} 
\multicolumn{1}{c|}{} & ChatGLM3 & \multicolumn{1}{r|}{{{9.43}}} & \multicolumn{1}{r|}{{{10.80}}} & {{11.51}} & \multicolumn{1}{r|}{{{10.47}}} & \multicolumn{1}{r|}{{{5.47}}} & {{6.64}} \\ \cline{2-8} 
\multicolumn{1}{c|}{} & Qwen & \multicolumn{1}{r|}{{{9.01}}} & \multicolumn{1}{r|}{\textit{\textbf{5.64}}} & {{6.07}} & \multicolumn{1}{r|}{{{9.96}}} & \multicolumn{1}{r|}{{{5.22}}} & {{6.16}} \\ 
\hline
\end{tabular}%
\caption{Results of Prompting, Finetuning and Multimodal Apporach
*Test Set of ASR-EC A, and B are respectively generated by Kaldi-K1, a DNN-HMM-Hybrid ASR system,  and Kaldi-K2, an end-to-end system which is based on the Zipformer-Transducer architecture}
\label{tab:results}
\end{table*}

We report the evaluation results of prompting and finetuning of LLMs in Table \ref{tab:results}. From the results, we emphasize the following key findings.

\begin{figure}[htp]
    \centering
    \includegraphics[width=0.6\linewidth]{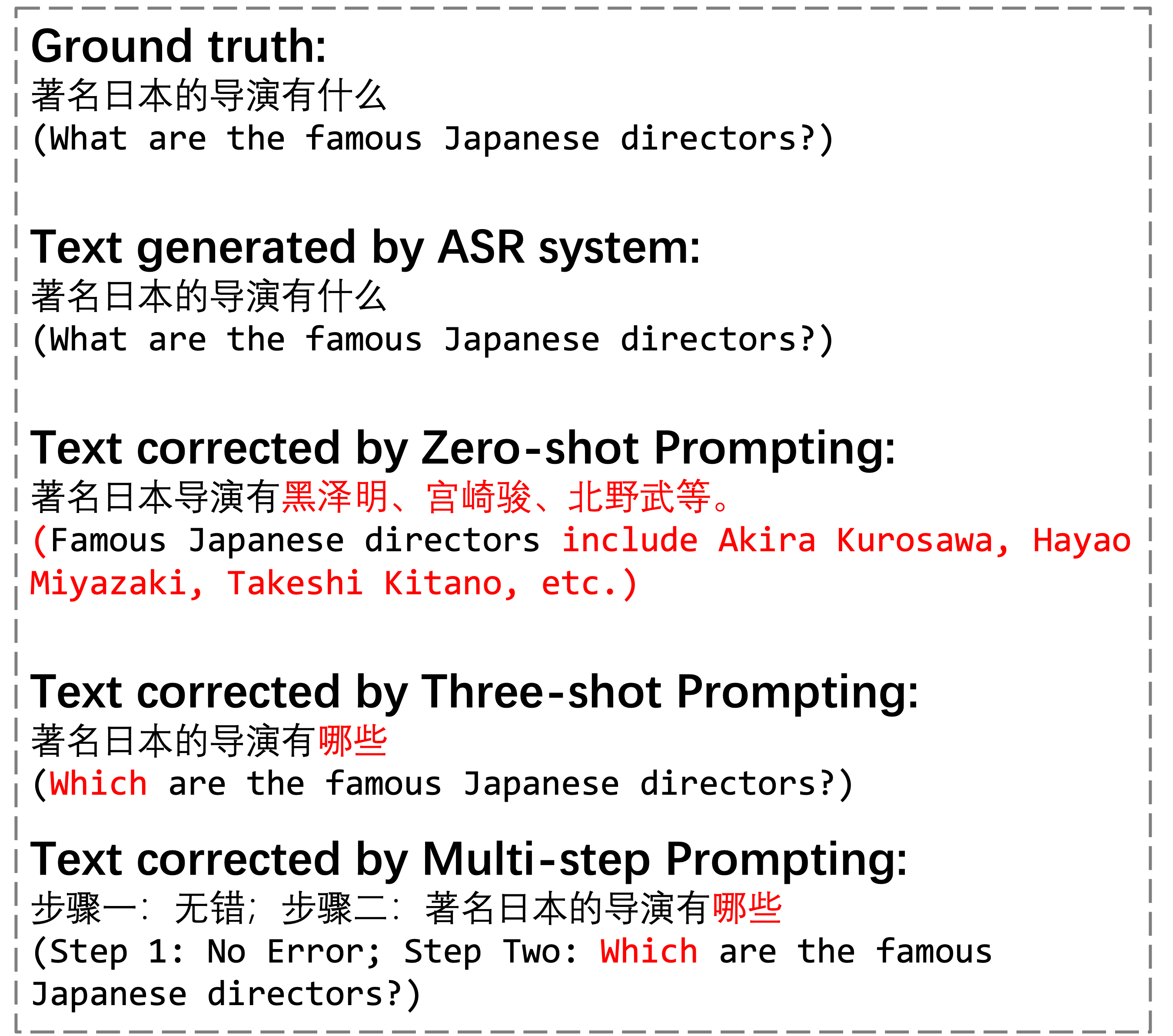}
    \caption{Outputs of Prompting Approaches}
    \label{fig:prompting}
\end{figure}

From Table~\ref{tab:results}, we observe that the prompting of LLMs had very poor performance, and their CER was significantly larger than that of ASR baselines and LoRA finetuning. 
Under the zero-shot and one-shot settings, LLMs tend to correct every sentence, regardless of whether the sentence has errors or not, and thus, the CER actually increased after correction by LLMs. 
The multi-step prompting method, which first lets the LLMs judge whether a sentence is correct and then corrects sentences labeled as incorrect, was found to mitigate the issue of over-correction in longer sentences. 
However, it showed no improvement for shorter sentences, likely due to the lack of contextual information necessary for LLMs to accurately complete the initial step of judging correctness. 
Because of the over-correction of LLMs, directly prompting the foundation models often fail to achieve good results. Specifically, the longer the sentence is corrected, the more over-correction the output has. 
Figure~\ref{fig:prompting} demonstrates several examples of these prompting approaches. 
This result also implies that LLMs can not achieve satisfactory performance without utilizing annotated ASR error correction datasets. 

\begin{figure}[htp]
  \centering
  \includegraphics[width=0.6\linewidth]{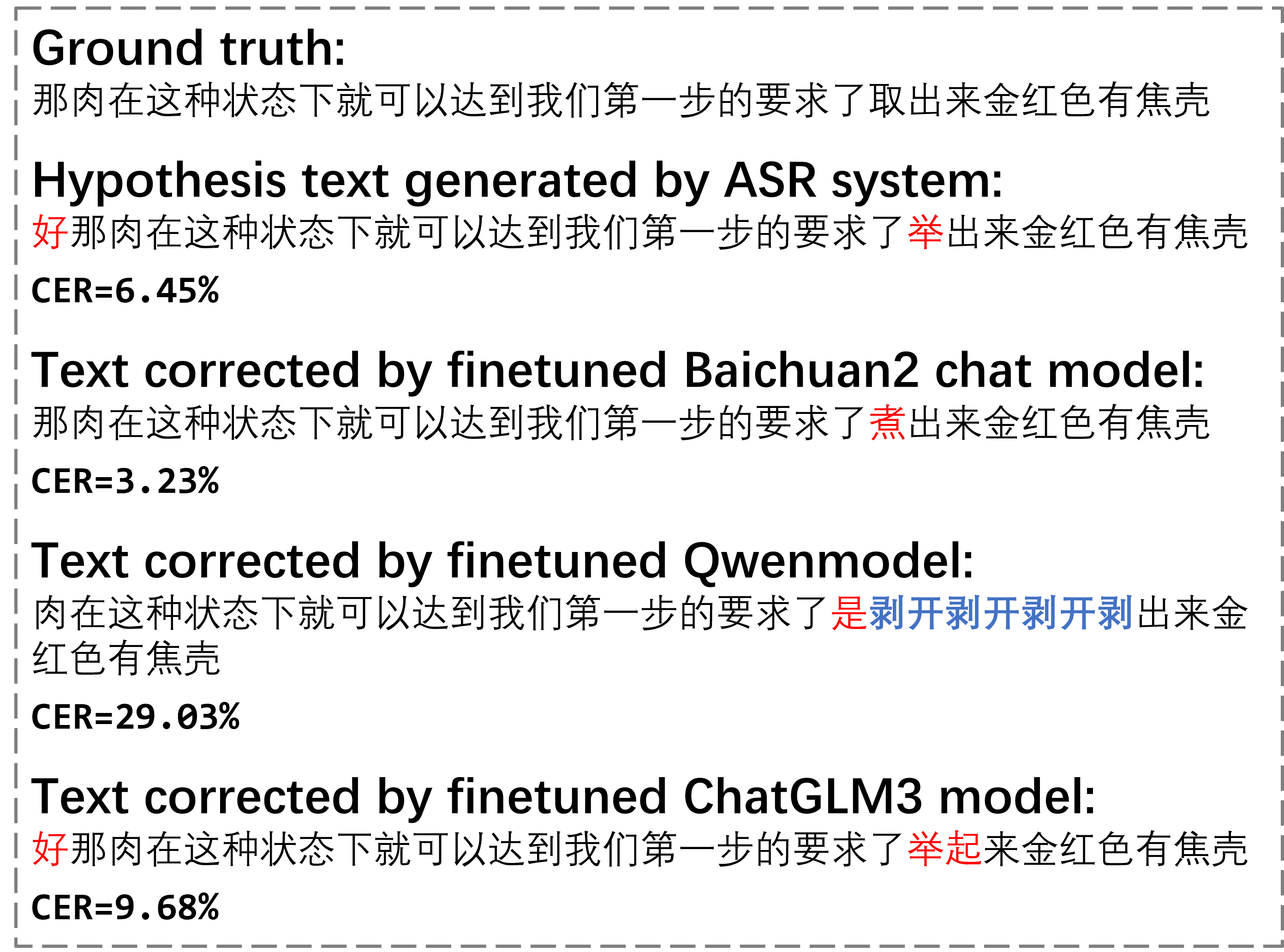}
  \caption{Outputs of different fine-tuned models.}
  \label{fig:model_output}
\end{figure}

By fine-tuning the open-source large models, the performance of all fine-tuned LLMs showed significant improvement, as revealed in Table~\ref{tab:results}. 
The findings reveal that open-source LLMs possess considerable capabilities when it comes to our benchmark. 
Our benchmark ASR-EC provides a supervised signal regarding the ASR error and highly resolves the over-correction problem in the prompting. 
The Baichuan2 chat model showed the best performance after fine-tuning and demonstrated the greatest improvement compared to its performance before fine-tuning. By analyzing the models' outputs, we found that the output of the Baichuan2 model is less likely to generate repeated text, while the outputs of Qwen and ChatGLM3 models show more repetition, which led to some correction failures, as shown in Figure~\ref{fig:model_output}. We speculate that this may be due to the Baichuan2 model's optimizer using max z loss during pre-training. This helped stabilize training and make the inference more robust to hyper-parameters.

\begin{figure*}[htp]
  \centering
  \includegraphics[width=1\linewidth]{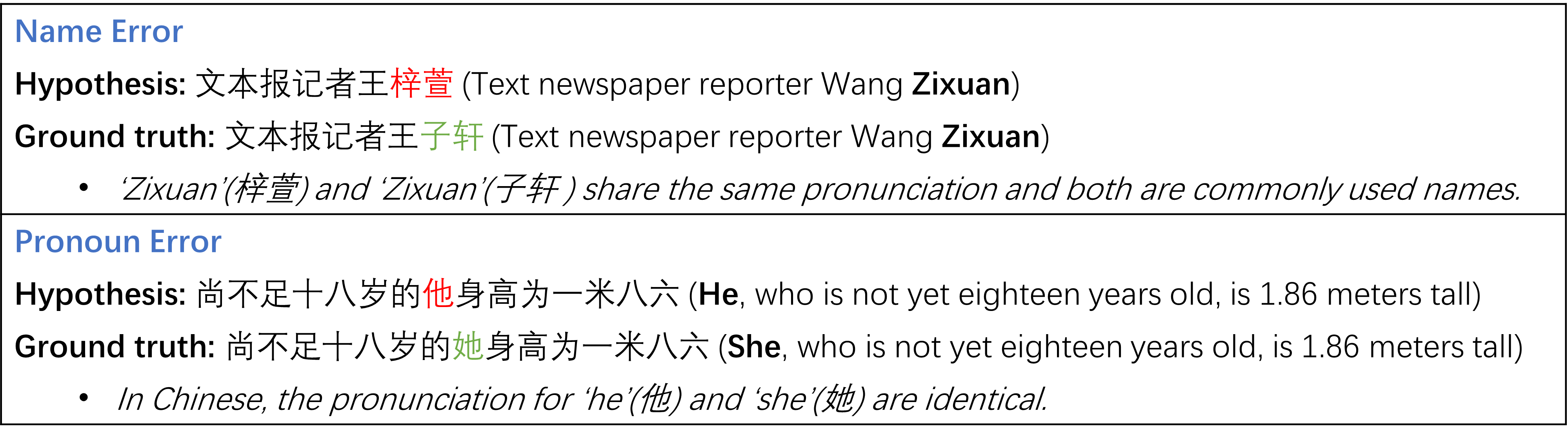}
  \caption{LLMs are not able to correct some types of errors without context or prior knowledge.}
  \label{fig:error_types}
\end{figure*}
\begin{table*}[htp]
\centering
\resizebox{\textwidth}{!}{%
\begin{tabular}{p{0.5cm}p{0.5cm}|rrr|rrr}
\hline
 & \multicolumn{1}{c|}{} & \multicolumn{3}{c|}{CER of Test Dataset A*} & \multicolumn{3}{c}{CER of Test Dataset B*} \\ \cline{3-8} 
 & \multicolumn{1}{c|}{} & \multicolumn{1}{c|}{\begin{tabular}[c]{@{}c@{}}Short \\ Utterances\end{tabular}} & \multicolumn{1}{c|}{\begin{tabular}[c]{@{}c@{}}Long \\ Utterances\end{tabular}} & \multicolumn{1}{c|}{\begin{tabular}[c]{@{}c@{}}Mixed\\  Utterances\end{tabular}} & \multicolumn{1}{c|}{\begin{tabular}[c]{@{}c@{}}Short \\ Utterances\end{tabular}} & \multicolumn{1}{c|}{\begin{tabular}[c]{@{}c@{}}Long \\ Utterances\end{tabular}} & \multicolumn{1}{c}{\begin{tabular}[c]{@{}c@{}}Mixed\\  Utterance\end{tabular}} \\ \hline
\multicolumn{2}{c|}{ASR Baselines} & \multicolumn{1}{r|}{\textit{\textbf{12.83}}} & \multicolumn{1}{r|}{\textit{\textbf{11.73}}} & \textit{\textbf{12.42}} & \multicolumn{1}{r|}{\textit{\textbf{13.44}}} & \multicolumn{1}{r|}{\textit{\textbf{6.92}}} & \textit{\textbf{8.11}} \\ \hline
\multicolumn{2}{c|}{Error Correction Baselines} & \multicolumn{1}{r|}{\textit{\textbf{8.36}}} & \multicolumn{1}{r|}{\textit{\textbf{24.40}}} & \textit{\textbf{22.58}} & \multicolumn{1}{r|}{\textit{\textbf{9.06}}} & \multicolumn{1}{r|}{\textit{\textbf{23.80}}} & \textit{\textbf{14.90}} \\ \hline 
\multicolumn{2}{c|}{Qwen-Audio ASR mode} & \multicolumn{1}{r|}{\textit{\textbf{3.82}}} & \multicolumn{1}{r|}{\textit{\textbf{4.06}}} & \textit{\textbf{5.58}} & \multicolumn{1}{r|}{\textit{\textbf{6.21}}} & \multicolumn{1}{r|}{\textit{\textbf{4.79}}} & \textit{\textbf{6.05}} \\ \hline
\multicolumn{2}{c|}{Qwen-Audio (Finetuning)} & \multicolumn{1}{r|}{\textit{\textbf{2.97}}} & \multicolumn{1}{r|}{\textit{\textbf{3.39}}} & \textit{\textbf{3.24}} & \multicolumn{1}{r|}{\textit{\textbf{4.35}}} & \multicolumn{1}{r|}{\textit{\textbf{3.52}}} & \textit{\textbf{3.66}} \\ \hline
\end{tabular}%
}
\caption{Enhancing Multimodal LLMs with ASR-EC
*Test Dataset A, and B are respectively generated by Kaldi-K1, a DNN-HMM-Hybrid ASR system,  and Kaldi-K2, an end-to-end system which is based on the Zipformer-Transducer architecture}
\label{tab:qwen-audio}
\end{table*}
\subsection{Multimodal Approach for LLMs}

We report the results of our multimodal LLM-based approach for ASR error correction and the comparison with baselines in Table~\ref{tab:results}. 
Compared with ASR baselines and the prompting and finetuning of LLMs, the multimodal approach achieves significant improvement.  
The integration of multimodal augmentation into ASR error correction capitalizes on the synergistic relationship between audio and text modalities, presenting a sophisticated approach to identifying and rectifying speech recognition errors. 
By employing a dynamic fusion process, multimodal augmentation combines the complementary strengths of audio signals and textual data. This fusion facilitates a comprehensive comprehension of the content, enabling the detection and rectification of errors that may not be readily apparent when analyzing each modality independently. Once errors are detected, multimodal augmentation applies contextually informed corrections, ensuring that the final text accurately captures the intended speech content. This methodology proves particularly effective in scenarios where traditional single-modality error correction techniques may prove insufficient.

\subsection{Enhancing Multimodal LLMs with Our ASR-EC Benchmark}

To further demonstrate the value and significance of our proposed benchmark ASR-EC for Chinese ASR error correction, we conducted experiments on the multimodal LLM, namely \emph{Qwen-Audio}, which involves both the modals of audio and text. 
The information of Qwen-Audio is shown in Table~\ref{tab:models}. 
Table~\ref{tab:qwen-audio} demonstrates the comparison of ASR baseline, the Qwen-audio-chat ASR mode (which simply performs the ASR without utilizing our ASR-EC dataset) and the combination of ASR and the Qwen-Audio finetuned with our ASR-EC dataset. 
As the table reveals, the latter two models significantly outperform the ASR baseline, and compared with the vanilla Qwen-Audio, the model finetuned with the ASR-EC dataset achieved a notable improvement on the CER. These results demonstrate that even for the multimodal LLMs, which are pre-trained on both audio and text datasets, our ASR-EC can still be significantly beneficial.

Furthermore, some errors in the hypothesis texts outputted by the ASR system can not be corrected by LLMs due to the lack of context and prior knowledge. Therefore, there is a minimum threshold CER that LLMs' error correction cannot surpass. We have qualitatively divided these errors into two categories: name errors and pronoun errors, with specific examples of each type of error provided in Figure \ref{fig:error_types}. 
Thus, we believe that our LLM-based multimodal approach and Qwen-M finetuned with our dataset have already achieved nearly optimal performances in the scenarios where the context and prior knowledge is missing.

\section{Conclusion}
\label{sec:con}

In conclusion, our research demonstrates a significant advancement in integrating Large Language Models (LLMs) for enhancing Chinese Automatic Speech Recognition (ASR) systems through Error Correction (EC). Our approach involved developing a specialized ASR-EC benchmark and applying methods such as parameter-efficient fine-tuning (PEFT) and multi-step prompting. Our experimentation reveals that while multi-step prompting reduces overcorrection issues inherent in direct prompting methods, fine-tuning with techniques like LoRA is crucial for significantly enhancing model performance. Despite these advancements, LLMs still face challenges in correcting errors requiring deep contextual understanding.

\bibliographystyle{IEEEtran}
\bibliography{IEEEabrv,reference}




\end{document}